\documentclass[11pt,a4paper]{article}
\usepackage[hyperref]{acl2021}
\usepackage{times}
\usepackage{latexsym}

\usepackage{bm}
\usepackage{tabularx}
\usepackage{amsmath}
\usepackage{amssymb}
\usepackage{booktabs}
\usepackage{multirow}
\usepackage{multicol}
\usepackage{balance}
\usepackage[flushleft]{threeparttable}
\usepackage{graphicx}
\usepackage{url}
\usepackage{microtype}

\aclfinalcopy % Uncomment this line for the final submission
\title{Assessing Dialogue Systems with Distribution Distances}

\author{Jiannan Xiang$^{1}$\footnotemark[1] \quad Yahui Liu$^{2}$\thanks{~~Equal contribution. Work was done during internship at Tencent AI Lab.} \quad Deng Cai$^3$ \quad Huayang Li$^4$ \quad Defu Lian$^1$ \quad Lemao Liu$^4$ \\
  $^1$University of Science and Technology of China \quad $^2$University of Trento, Italy \\
  $^3$The Chinese University of Hong Kong \quad $^4$Tencent AI Lab, China \\
  }

\date{}

\begin{document}
\maketitle

\begin{abstract}
An important aspect of developing dialogue systems is how to evaluate and compare the performance of different systems. Existing automatic evaluation metrics are based on turn-level quality evaluation and use average scores for system-level comparison. In this paper, we propose to measure the performance of a dialogue system by computing the distribution-wise distance between its generated conversations and real-world conversations. Specifically, two distribution-wise metrics, FBD and PRD, are developed and evaluated. Experiments on several dialogue corpora show that our proposed metrics correlate better with human judgments than existing metrics.
\end{abstract}

\section{Introduction}
Dialogue generation is a special text generation task, which has drawn booming attention in the natural language processing community. It is widely agreed that one single input query is often associated with multiple valid responses in this task, which is termed as a $1$-to-$n$ relationship between a query and its responses~\cite{vinyals2015neural,zhou2017mechanism,zhao-etal-2017-learning,liu2018towards,wang2021a,chan2021enhancing,gao2021ream}. It increases the challenges of automatically evaluating the performance of dialogue systems. 

In general, the previous evaluation metrics mainly focus on turn-level quality. For example, unsupervised word-overlapping or embedding-based metrics~\cite{Papineni2002BleuAM,Lin2004ROUGEAP,Mitchell2008VectorbasedMO,zhang2019bertscore} calculate the similarity or alignment between generated responses and reference responses, which is not well-suited for open-end dialogue tasks. Learned classification or regression systems~\cite{Lowe2017TowardsAA,tao2018ruber,sellam2020bleurt,Ghazarian2019BetterAE} are corpus-dependent because of requiring additional task-specific training or tuning, which run the risk of assigning lower quality to a better model in the overfitting or underfitting cases. 

%\jnxiang{Do we need to detail previous works so much?}Currently, most automatic evaluation metrics in this field are dominated by turn-level quality measure: (1) Inspired by the practice in machine translation~\cite{papineni2002bleu} and text summarization~\cite{lin2004rouge}, the similarity score between the generated response and the reference response is used~\cite{zhang2019bertscore,sellam2020bleurt}.
%This strategy requires the generated sentences to strictly align with the golden references, which is not suitable for open-end dialogue tasks. 
%(2) Training a classification or regression system for quality scoring between context and response~\cite{sellam2020bleurt,tao2018ruber}. This strategy necessitates training/tuning of a classifier for each corpus which is seldom practical. Furthermore, the classifier might detect a single dimension where the true and generated samples differ and enjoy high accuracy, which runs the risk of assigning lower quality to a better model. 

%In general, the previous evaluation metrics are mainly focus on turn-level quality. 
In this paper, we provide a new perspective that distribution distance between generated conversations and real conversations can be applied to measure the performances of dialogue systems. There are three main \textit{contributions}: (1) We firstly propose two unsupervised distribution-wise metrics (i.e., FBD and PRD) to solve the evaluation issue in this field. (2) The experimental results show that the proposed distribution-wise metrics perform well. Particularly, FBD achieves compelling performances on most evaluation corpora, which shows a promising direction for designing evaluation metrics. (3) We collect the typical evaluation corpora and existing evaluation metrics in order to better assess the performance of dialogue systems, which could be useful for researchers in this community~\footnote{The source codes and data are available at \url{https://github.com/yhlleo/frechet-bert-distance}.}.
%\jnxiang{Briefly describe our experimental results here?}

\section{Related Work}
In this section, we focus on \textit{unsupervised} automatic evaluation metrics for dialogue system evaluation. In general, existing unsupervised metrics mainly measure \textit{turn-level} qualities, which can be categorized into two main classes: word overlapping metrics and embedding-based metrics:

\noindent \textbf{Word-overlapping Metrics}
  Such metrics quantify the amount of word-overlap between generated response and reference responses. %\jnxiang{This sentence is a little confusing.} One way is to base the comparison on word overlap. 
  For example, BLEU~\citep{Papineni2002BleuAM} calculates the geometric mean of the precision for $n$-gram. ROUGE~\citep{Lin2004ROUGEAP} is a recall-oriented metric. METEOR~\citep{Banerjee2005METEORAA} computes the harmonic mean of precision and recall
  %and improves $n$-gram matching 
  with stemming and synonyms.
 
\noindent \textbf{Embedding-based Metrics}
  Embedding-based metrics align the generated response and the reference in latent semantic space. Some adopt the vector similarity of sentence embeddings as a quality measure. For example, Embedding Average~\citep{Foltz1998TheMO, Mitchell2008VectorbasedMO} calculates sentence-level embeddings by averaging word representations. Vector Extrema~\citep{forgues2014bootstrapping} computes sentence-level embeddings by taking the most extreme value for each dimension in all word vectors. Others adopt more fine-grained semantic matching. For example, Greedy Matching~\citep{rus2012optimal} greedily matches each word in a generated response to a word in the reference response, and the final score is defined as the average of word-level similarity scores. %\citet{Lowe2017TowardsAA} proposed to learn a neural model that predicts the score of a response given the context and a reference response. 
  \citet{zhang2019bertscore} introduced a better embedding-based metric BERTScore that computes word similarity using contextual embeddings from pre-trained language models.
%based on the similarity of word embeddings
%\jnxiang{Not emphasize the shortcoming of reference-based metrics here? (since our method is also refer-based)}The aforementioned metrics rely on reference responses, which could be problematic due to the one-to-many nature of dialogue~\cite{zhao-etal-2017-learning,gupta2019investigating}. Several reference-free metrics have been explored for dialog response evaluation \cite{li-etal-2017-adversarial,Tao2018RUBERAU,Ghazarian2019BetterAE}. 
%Most recently, \citet{Mehri2020USRAU} proposed USR, a reference-free metric composed of multiple sub-metrics. However, these reference-free methods often rely on task-specific training with task-specific data.

Our proposed methods are best placed in the literature of embedding-based metrics. However, there are two main differences from previous metrics in this field: (1) We compute the distribution distance between embedding sets as the system-level performance of a dialogue system, which does not require task-specific training/tuning; (2) We propose to extract sentence-level semantic representations directly from pre-trained language models~\cite{jacob2018bert,liu2019roberta}, where there are no operations of converting the wold-level embeddings to sentence-level embeddings.
%\yahui{differences between our methods and existing embedding-based methods.}

\section{Proposed Methods}

Given a collect of sentence pairs $\{(\bm{x}_i, \bm{y}_i)\}^N$, we assume that 
the corresponding semantic representations $\{\bm{v}_i\}^N$ can be extracted in this manner:
\begin{equation}
    \bm{v}_i = LM([\bm{x}_i, \bm{y}_i])
\end{equation}
where $LM(\cdot)$ refers to pre-trained language models(i.e., \cite{jacob2018bert,liu2019roberta,yang2019xlnet,clark2020electra}), $[\cdot,\cdot]$ refers to the concatenation operation. Intuitively, the differences between the distribution $R$ of real samples and the distribution $G$ of generated samples can be applied to measure the performances of dialogue systems (i.e., $d(R, G)$). Therefore,  we propose two distribution-based methods to automatically evaluate the performance of dialogue systems, which are presented in this section in detail. 

\subsection{Fr\'echet Bert Distance}
Semantic representations $\{\bm{v}_i\}^N$ are extracted by a pre-trained language model, which encodes the contextual information of the sentences. The main intuition is that the distribution of semantic representations of generated sentences should be as close as possible to the distribution of semantic representations of real sentences in a successful system. To measure this, we assume that such semantic representations follow a multi-dimensional Gaussian, which can be represented by variables: \textit{mean} and \textit{covariance}. The difference between two Gaussians (generated and real sentence pairs) is measured by the Fr\'echet distance~\cite{dowson1982frechet}. We call the Fr\'echet distance between the distribution $R$ with mean $(\bm{\mu}_r, \bm{\Sigma}_r)$ obtained from real sentence pairs and the distribution $G$ with mean $(\bm{\mu}_g, \bm{\Sigma}_g)$ obtained from generated sentence pairs as ``Fr\'echet Bert Distance" (FBD), which is formulated as:
\begin{equation}
\begin{split}
    d_{\mathrm{FBD}}( & R, G) = \|\bm{\mu}_r - \bm{\mu}_g\| + \\ & \mathrm{Tr}(\bm{\Sigma}_r + \bm{\Sigma}_g - 2 (\bm{\Sigma}_r\bm{\Sigma}_g)^{1/2}) \\
\end{split}
\end{equation}
Once the distribution of generated data closes to the distribution of real data, the model indeed achieves low FBD scores.
Similarly, such distance~\cite{heusel2017gans} has been widely verified in various Generative Adversarial Networks (GANs) in computer vision tasks~\cite{karras2017progressive,zhang2018stackgan++,park2019semantic}, which is consistent with increasing disturbances and human judgment. Surprisingly, we observed that FBD works well in evaluating open-end dialogue systems. 

\subsection{Precision-Recall Distance}

We notice that FBD is based on the estimated Gaussian parameters $(\bm{\mu}, \bm{\Sigma})$. There is an optional strategy to get rid of estimating the parameters.
%However, the assumption of multivariate Gaussian distribution sometimes could be a limitation to represent all the real distribution. 
Inspired by~\cite{sajjadi2018assessing}, we apply a precision-recall-based method, named as Precision-Recall Distance (PRD), to evaluate the distance between two distributions. 

The key intuition is that precision should measure how much of $G$ can be generated by a “part” of $R$ while recall should measure how much of $R$ can be generated by a “part”
of $G$. In general, (a) If $R$ is bimodal and $G$ only captures one of the modes, we should have perfect precision but only limited recall; (b) In
the opposite case, we should have perfect recall but only limited precision; (c) If $R$ = $G$, we should
have perfect precision and recall; (d) If the supports of $R$ and $G$ are disjoint, we should have zero
precision and recall. The BPD is formulated as:
\begin{equation}
    d_{\mathrm{PRD}}(R, G) = \max \bigg\{\frac{2\alpha(\lambda)\beta(\lambda)}{\alpha(\lambda) + \beta(\lambda)} \bigg| \lambda \in \Lambda\ \bigg\}
\end{equation}
where $\Lambda = \{\tan(\frac{i}{m+1}\frac{\pi}{2}) | i=1,\cdots, m\}$, $m\in\mathbb{N}$ refers to a given angular resolution, $\alpha(\lambda) = \sum_{\bm{v}\in\mathcal{V}} \min (\lambda R(\bm{v}), G(\bm{v}))$ and $\beta(\lambda) =\sum_{\bm{v}\in\mathcal{V}}\min(R(\bm{v}), \frac{G(\bm{v})}{\lambda})$~\footnote{For a distribution $P$ with a finite state space $\mathcal{V}$, we have $\bm{v} \in \mathcal{V}$ and $P(\bm{v}) > 0$.}. Therefore, the better dialogue systems will achieve higher PRD scores. 

\section{Experiments}

\subsection{Datasets \& Systems}
To verify the two proposed metrics, we conduct experiments on six public dialogue corpora. 

\noindent\textbf{Baseline Metrics}. We mainly compare with several widely-used metrics in text generation field: a) three word-overlapping metrics:  BLEU~\cite{Papineni2002BleuAM}, ROUGE~\cite{Lin2004ROUGEAP}, METEOR~\cite{denkowski2014meteor}; b) four embedding-based metrics: Greedy Matching~\cite{rus2012optimal}, Embedding Average~\cite{wieting2015towards}, Vector Extrema~\cite{forgues2014bootstrapping} and BERTScore~\cite{zhang2019bertscore}. All these metrics do not require task-specific training. 
%\cd{ $\Rightarrow$ do not require task-specific training (note some other metrics such as USR also do not need annotated corpus. maybe use an additional table for the comparison with USR.)}

\noindent\textbf{Datasets}. We collect three recently released evaluation corpora which consist of dialogue query and response samples of different systems, and the corresponding human annotations:

\begin{table}[!ht]
    \renewcommand{\tabcolsep}{3pt}
    \renewcommand{\arraystretch}{1.2}
    \small
	\centering
	\begin{tabularx}{\columnwidth}{l|p{5.8cm}}
	\toprule
	\textbf{Source} & \textbf{Address} \\ \hline
	Transformers & \url{https://github.com/huggingface/transformers} \\ \hline
	USR & \url{https://github.com/Shikib/usr} \\ \hline
	GRADE & \url{https://github.com/li3cmz/GRADE/tree/main/evaluation} \\ \hline
	Daily(Z) & \url{https://github.com/ZHAOTING/dialog-processing/tree/master/src/tasks/response_eval} \\ 
	\multirow{-4}{*}{Persona(Z)} & \\ 
	\hline 
	ParlAI & \url{https://github.com/facebookresearch/ParlAI} \\ \hline
	BLEU & \multirow{2}{*}{\url{https://github.com/nltk/nltk}} \\ 
	METEOR & \\ \hline
	ROUGE-L & \\
	Greedy & \url{https://github.com/Maluuba/nlg-eval} \\ 
	\multirow{-3}{*}{Average} & \\
	\multirow{-3}{*}{Extrema} \\ \hline
	BERTScore & \url{https://github.com/Tiiiger/bert_score} \\
     \bottomrule
	\end{tabularx}
	\caption{\label{tab:sources-table} All the public resources in our experiments.
	}
\end{table} 

\begin{table}[!ht]
    \renewcommand{\tabcolsep}{2pt}
    \renewcommand{\arraystretch}{1.2}
    \small
	\centering
	\begin{tabular}{l|l|l}
	\toprule
	\multirow{2}{*}{\textbf{Corpus}} & \textbf{Num. of } & \multirow{2}{*}{\textbf{Systems}} \\ 
	& \textbf{Samples} & \\ \hline
	\multirow{4}{*}{Persona(M)} & \multirow{4}{*}{60} & Seq2Seq \\ 
	& & LSTM language model \\
	& & Key-Value Profile Memory Network \\
	& & Generated Human-written \\ \hline
	Daily(H) & 150 & \multirow{2}{*}{Transformer-Ranker} \\
	Convai2 & 150 & \multirow{2}{*}{Transformer-Generator} \\
	Empathetic & 150 &  \\ \hline
	\multirow{5}{*}{Daily(Z)} & \multirow{5}{*}{100} & Seq2Seq \\
	\multirow{5}{*}{Persona(Z)} & \multirow{5}{*}{150} & Attentional Seq2Seq   \\
	& & HRED \\
	& & VHRED \\
	& & GPT2-sm \\
	& & GPT2-md \\
	\bottomrule
	\end{tabular}
	\caption{\label{tab:corpus-details-table} The details of each evaluation corpus.
	}
\end{table} 

\begin{itemize}
    %\vspace{-0.5em}
    \item \textbf{Persona(M)}: USR~\cite{Mehri2020USRAU} built an evaluation corpus based on PersonaChat~\cite{zhang2018personalizing}, in which both four system outputs and the corresponding human evaluation scores were collected.
    %(i.e., understandable, natural, maintains context, interesting, uses knowledge and overall quality) were well annotated.
    %\vspace{-0.9em}
    \item \textbf{Daily(H), Convai2, and Empathetic}: GRADE~\cite{huang2020grade} used three dialogue corpora, including DailyDialog~\cite{Lowe2017TowardsAA}, Convai2~\cite{dinan2019second} and EmpatheticDialogues~\cite{rashkin2018towards}, to do the evaluations and compared two dialogue models: Transformer-Ranker and Transformer-Generator collected from the ParlAI platform~\cite{miller2017parlai}.
    %\vspace{-0.9em}
    \item \textbf{Daily(Z) and Persona(Z)}: Dialogue Evaluation~\cite{zhao2020designing} used DailyDialog~\cite{Lowe2017TowardsAA} and PersonaChat~\cite{zhang2018personalizing} to build two evaluation corpora and collected outputs of six generative models (with three decoding strategies). The appropriateness of each response was obtained by human annotation.
\end{itemize}
%\cd{how about Persona(M), Daily(H), Convai2, Empathetic, Daily(Z) Persona(Z)}

\begin{table*}[!ht]
\centering
\renewcommand{\arraystretch}{0.9}
\begin{tabularx}{\linewidth}{lrr rr rr rr rr rr}
\toprule 
\multirow{2}{*}{\textbf{Metric}} & \multicolumn{2}{c}{\textbf{Persona(M)}} &
\multicolumn{2}{c}{\textbf{Daily(H)}} &
\multicolumn{2}{c}{\textbf{Convai2}} & \multicolumn{2}{c}{\textbf{Empathetic}} & \multicolumn{2}{c}{\textbf{Daily(Z)}} & 
\multicolumn{2}{c}{\textbf{Persona(Z)}} \\ \cmidrule(lr){2-3} \cmidrule(lr){4-5} \cmidrule(lr){6-7} \cmidrule(lr){8-9} \cmidrule(lr){10-11} \cmidrule(lr){12-13}
& Spr. & Pr. & Spr. & Pr.  & Spr. & Pr.  & Spr. & Pr.  & Spr. & Pr. & Spr. & Pr.  \\ \hline
\multicolumn{13}{c}{Word-Overlapping Metrics} \\ \hline
BLEU & .400 & .672 & .445 & .444 & .800 & .801 & .136 & .331 & .595 & .421 & .400 & .390  \\
METEOR & .800 & .860 &  .018 & .050 & .800 & .767 & .382 & .133 & .643 & .689 & .700 & .936  \\
ROUGE-L & .600 & .289  & .545 & .417 & .200 & .061 & .391 & .472 & .738 & .725 & .400 & .915 \\ \hline
\multicolumn{13}{c}{Embedding-Based Metrics} \\ \hline
Greedy & .600 & .260 &  .855 & .764 & .600 & .794 & .736 & .864 & .690 & .726 & .100 & .835 \\
Average & .800 & \textbf{.863} & .209 & .209 & .600 & .879 & .664 & .725 & .548 & .769 & .300 & .861 \\
Extrema & .600 & .435 & .745 & .761 & .800 & .766 & .618 & .722 & .595 & .746 & .500 & .834 \\
BERTScore$^{\text{B}}$ & .800 & .590 &  .137 & .082 & .800 & .817 & .300 & .077 & .857 & .883 & .900 & .961  \\ 
BERTScore$^{\text{R}}$ & .800 & .517 & .855 & .857 & .800 & .939 & .600 & .697 & .714 & .860 & .700 & .918 \\ \hline
%\multicolumn{13}{c}{Reference Free Metric}  \\ \hline
%USR & \textbf{1.00} & .820 & - & - & - & - & - & - & - & - & - & -  \\ \hline 
\multicolumn{13}{c}{Distribution-Based Metrics} \\ \hline
FBD$^{\text{B}}$ & \textbf{1.00}  & .853 & .564 & .717 & .800 & .854 & .427 & .623 & .786 & .763 & .400 & .923 \\
FBD$^{\text{R}}$ & \textbf{1.00} & .802 & \textbf{.891} & \textbf{.926} & .800 & .747 & \textbf{.864} & \textbf{.951} & \textbf{.929} & \textbf{.963}  & \textbf{1.00} & .860 \\
PRD$^{\text{B}}$ & .800  & .637 & .221 & .409 & \textbf{1.00} & \textbf{.972} & .227 & .399 &  .690 & .914 & .900 & \textbf{.984}  \\ 
PRD$^{\text{R}}$ & .800 & .660 & .591 & .578 & \textbf{1.00} & .913 & .545 & .583 & .762 & .906 &  .900 & .932 \\ \bottomrule
\end{tabularx}
\begin{tablenotes}
    \small
    \item 1.``Spr." and ``Pr." refer to Spearman and Pearson correlation coefficients, respectively.
    \item 2. $^{\text{B}}$ and $^{\text{R}}$ mean using BERT (base) and RoBERTa (base) as language models, respectively.
\end{tablenotes}
%\vspace{-0.6em}
\caption{\label{overall-table} Correlations of all the metrics with overall quality ratings the six dialogue corpora.}
%\vspace{-1.2em}
\end{table*}

\begin{figure}[htbp]	
%\vspace{-0.2em}
	\centering
	\begin{tabular}{c}
		\includegraphics[width=\linewidth]{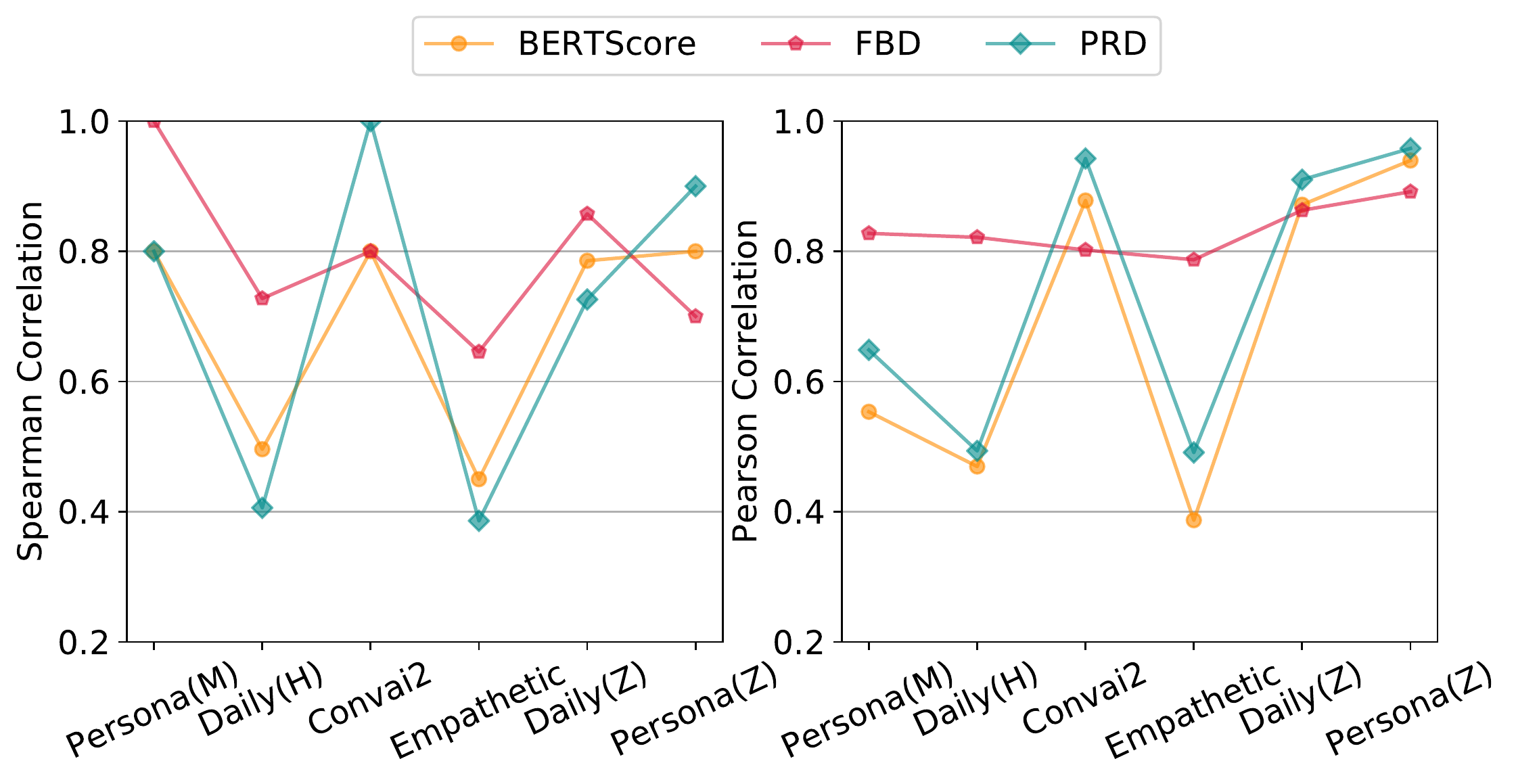} \\
	\end{tabular}
	%\vspace{-1.2em}
    \caption{\label{fig:corpus-metrics} The comparisons between BERTScore, FBD and PRD metrics on various corpora. For each corpus, we average the performances of  BERT and RoBERTa language models.}
    %\vspace{-1.2em}
\end{figure}

%\vspace{-0.45em}
\noindent\textbf{Implementation Details}. We leverage pre-trained BERT~\cite{jacob2018bert} and RoBERTa~\cite{liu2019roberta} for utterance-level contextualized encoding~\footnote{Based on the public project: \url{https://github.com/huggingface/transformers}} without additional tuning or training the language models. For each query and response pair $(\bm{x}_i, \bm{y}_i)$, we use the last hidden output of $[\textsc{CLS}]$ as its semantics representation without tuning or training the language models. 
%The pre-trained BERT and RoBERTa language models are directly downloaded from the open-source project without any additional training or tuning. 
To assess the system-level performances of dialogue systems, we calculate the Spearman and Pearson correlations between the rankings of human evaluation and the rankings of evaluation metrics. If a evaluation metric is designed for \textit{turn-level} evaluation, we average the all turn-level scores as the performance of the corresponding dialogue system. 

\begin{figure}[htbp]	
	\centering
	\begin{tabular}{c}
		\includegraphics[width=\linewidth]{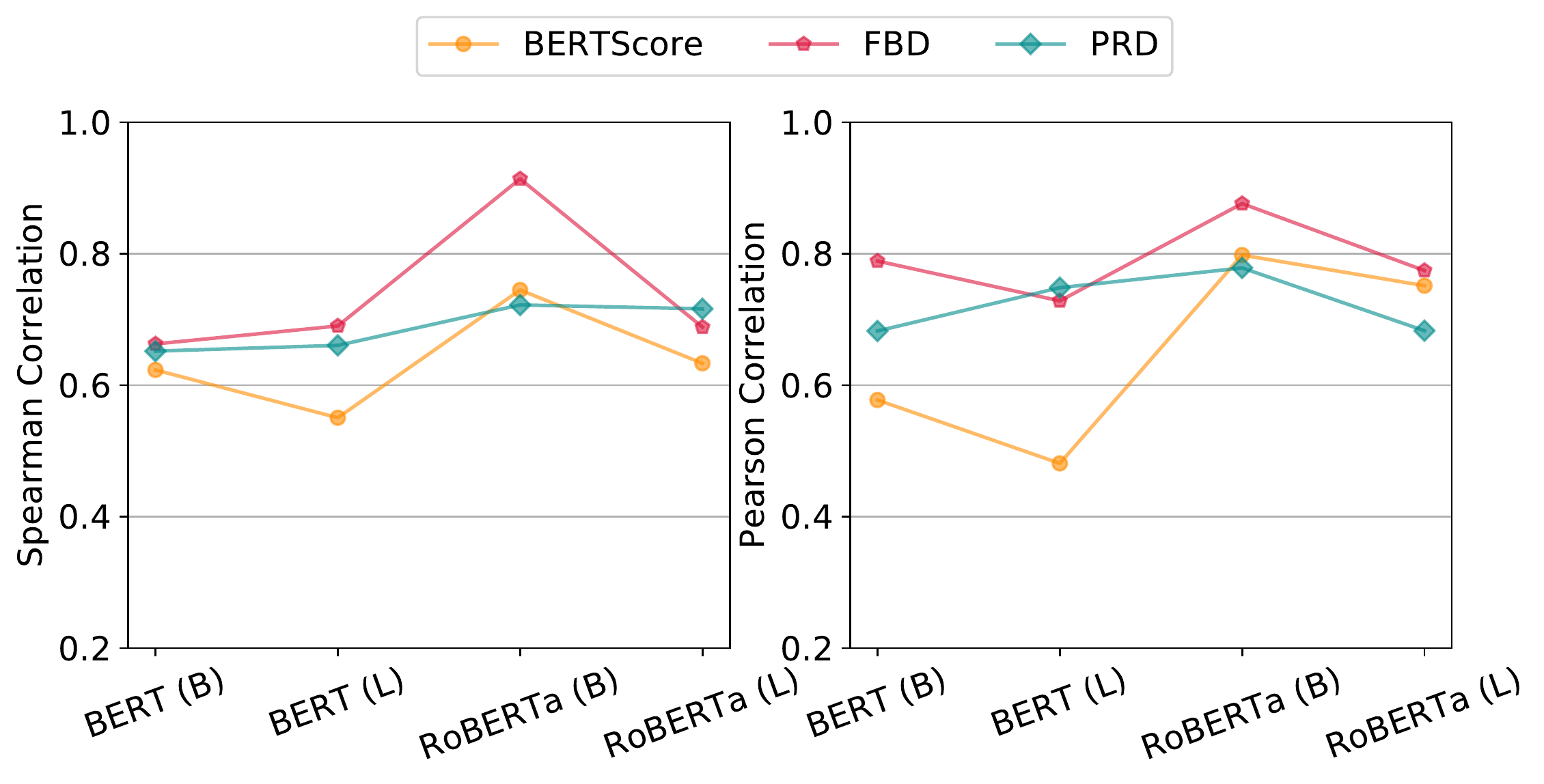} \\
	\end{tabular}
	%\vspace{-1.2em}
    \caption{\label{fig:models-metrics} Comparisons between BERTScore, FBD and PRD metrics on various language models. For each model, we average the performances on all the six evaluation corpora.}
\end{figure}

\noindent\textbf{Public Resources}. All the compared evaluation corpora and evaluation metrics are available in Table~\ref{tab:sources-table}. Once the official implementations are not available, we use the repositories with highest ``stars" on GitHub. The details of each evaluation corpus, including number of samples and compared dialogue systems in each corpus, are presented in Table~\ref{tab:corpus-details-table}.

% \begin{table}[!ht]
%     \renewcommand{\tabcolsep}{3pt}
%     \renewcommand{\arraystretch}{1.2}
%     \small
% 	\centering
% 	\begin{tabular}{l|l|l|l}
% 	\toprule
% 	\multirow{2}{*}{\textbf{Corpus}} & \textbf{Num. of } & \multirow{2}{*}{\textbf{Systems}} & \textbf{Decoding} \\ 
% 	& \textbf{Samples} & & \textbf{Strategy}\\ \hline
% 	\multirow{4}{*}{Persona(M)} & \multirow{4}{*}{60} & Seq2Seq & \multirow{4}{*}{Default} \\ 
% 	& & LSTM language model \\
% 	& & Key-Value Profile Memory Network \\
% 	& & Generated Human-written \\ \hline
% 	Daily(H) & 150 & \multirow{2}{*}{Transformer-Ranker} & \multirow{2}{*}{Default}\\
% 	Convai2 & 150 & \multirow{2}{*}{Transformer-Generator} \\
% 	Empathetic & 150 &  \\ \hline
% 	\multirow{5}{*}{Daily(Z)} & \multirow{5}{*}{100} & Seq2Seq & \multirow{3}{*}{Greedy Decoding} \\
% 	\multirow{5}{*}{Persona(Z)} & \multirow{5}{*}{150} & Attentional Seq2Seq & \multirow{3}{*}{Ancestral Sampling}  \\
% 	& & HRED  & \multirow{3}{*}{Nucleus Sampling}\\
% 	& & VHRED \\
% 	& & GPT2-sm \\
% 	& & GPT2-md \\
% 	\bottomrule
% 	\end{tabular}
% 	\caption{\label{tab:corpus-details-table} The details of each evaluation corpus.
% 	}
% \end{table} 

\subsection{Results}
%\jnxiang{split into two sections?}
We compute the system-level correlation between all automatic metrics and the quality ratings by using Spearman and Pearson correlation coefficients.

\begin{table}[!ht]
\centering
\renewcommand{\arraystretch}{1.1}
\renewcommand{\tabcolsep}{2.3pt}
\small
\begin{tabularx}{\columnwidth}{l r r r r r r}
\toprule 
Model & P(M) & D(H) & C2 & EM & D(Z) & P(Z) \\ \hline
BERT &  .57\tiny{$\pm$.44} & .65\tiny{$\pm$.39}  & .60\tiny{$\pm$.42}  & .65\tiny{$\pm$.39} &  .67\tiny{$\pm$.38} & .65\tiny{$\pm$.38} \\
RoBERTa &  .71\tiny{$\pm$.38} & .68\tiny{$\pm$.37}  & .73\tiny{$\pm$.34}  &  .68\tiny{$\pm$.37} & .68\tiny{$\pm$.37} & .71\tiny{$\pm$.33} \\
%& B (B) & B (L) & R (B) &  R (L) \\  \hline
%Persona(M) & .57\tiny{$\pm$.44} & .42\tiny{$\pm$.42} & .71\tiny{$\pm$.38} & .71\tiny{$\pm$.35} \\
%Daily(H) & .65\tiny{$\pm$.39} & .49\tiny{$\pm$.45} & .68\tiny{$\pm$.37} & .70\tiny{$\pm$.36}\\
%Convai2 & .60\tiny{$\pm$.42} & .45\tiny{$\pm$.44} & .73\tiny{$\pm$.34} & .69\tiny{$\pm$.37}\\
%Empathetic & .65\tiny{$\pm$.39} & .49\tiny{$\pm$.45} & .68\tiny{$\pm$.37} & .70\tiny{$\pm$.36}\\
%Daily(Z) & .67\tiny{$\pm$.38} & .47\tiny{$\pm$.44} & .68\tiny{$\pm$.37} & .39\tiny{$\pm$.39} \\
%Persona(Z) & .65\tiny{$\pm$.38} & .44\tiny{$\pm$.41} & .71\tiny{$\pm$.33} & .72\tiny{$\pm$.33}\\
\bottomrule
\end{tabularx}
\begin{tablenotes}
    \small
    \item 1. $\{$P(M), D(H), $\cdots$, P(Z)$\}$ refer to the six evaluation corpora.
    \item 2. The reported values are mean\tiny{$\pm$standard deviation}.
\end{tablenotes}
%\vspace{-0.2em}
\caption{\label{normality-table}  Comparisons of the normality on various evaluation corpora. The normality is calculated over each dimension of the extracted semantic representations. }
%\vspace{-1em}
\end{table}

The performances of various evaluation metrics on different public corpora are reported in Table~\ref{overall-table}. Our proposed two metrics (i.e., FBD and PRD) show comparable performances over the baseline metrics. Especially, FBD$^\text{R}$ achieves compelling performances on five corpora, which indicates a good ability of generalization and robustness on various corpora. In addition, most evaluation metrics are sensitive to the evaluation corpora. For example, BLEU performs well in \textbf{Convai2} but fails in \textbf{Empathetic}. Similarly, BERTScore$^\text{B}$ performs well in \textbf{Convai2} and \textbf{Persona(Z)} but fails in \textbf{Daily(H)} and \textbf{Empathetic}. It indicates that the selection of evaluation corpora has a great influence on assessing the performances of evaluation metrics. Hence, it's better to use multiple corpora to do the comparisons between metrics. Obviously, our proposed FBD$^\text{R}$ outperforms the existing evaluation metrics in the view of robustness.
%of a metric on various corpora should be considered in the comparisons.

In Figure~\ref{fig:corpus-metrics}, we compare the evaluation metrics in the perspective of various evaluation corpora, where the results of BERT and RoBERTa language models are averaged. It confirms the superiority of the FBD metric. Compared to the performance of USR~\cite{Mehri2020USRAU} (1.000/.820 on Persona(M)), a reference-free metric that relies on task-specific training/tuning with task-specific data, the performances of our proposed methods are comparable without any training/tuning. Therefore, we believe that it's a promising direction to explore distribution-wise metrics for assessing dialogue systems in this field.

As shown in Figure~\ref{fig:models-metrics}, we average the performances of each metric on all evaluation corpora. It shows that our proposed FBD has higher performance expectations that outperform BERTScore with different language models. 
The large models do not show improvements in average performance compared to the base models. 
In general, FBD metric achieves better Spearman and Pearson correlations compared to PRD. 
\begin{table}[!ht]
    \renewcommand{\tabcolsep}{3pt}
    \renewcommand{\arraystretch}{1.2}
    \small
	\centering
	\begin{tabularx}{\columnwidth}{@{}Xl rr rr rr rrX@{}}
	\toprule
	\multirow{2}{*}{\textbf{Model}} & \multicolumn{2}{c}{\textbf{Relevance}} & \multicolumn{2}{c}{\textbf{Grammar}} & \multicolumn{2}{c}{\textbf{Content}} & \multicolumn{2}{c}{\textbf{Overall}} \\ 
	\cmidrule(lr){2-3} \cmidrule(lr){4-5} \cmidrule(lr){6-7}  \cmidrule(lr){8-9}
	 & Spr. & Pr. & Spr. & Pr.  & Spr. & Pr.  & Spr. & Pr. \\ \hline
        \multicolumn{9}{c}{Word-Overlapping Metrics} \\ \hline
        BLEU & .857 & .454 & .167 & .213 & .143 & .270 & .595 &.421  \\
        METEOR & .810 & .736 & .119 & .044 & .024 & .349 & .643 & .689 \\
        ROUGE-L & .857 & .758 & .238 & .075 & .190 & .375 & .738 & .725 \\ \hline
        \multicolumn{9}{c}{Embedding-Based Metrics} \\ \hline
        Greedy & .881 & .769 & .214 & .133 & .119 & .350 & .690 & .726 \\
        Average & .762 & .808 & .071 & .197 & .238 & .399 & .548 & .769  \\
        Extrema & .714 & .776 & .143 & .009 & .476 & .539 & .595 & .746 \\
        BERTScore$^{\text{B}}$ & \textbf{.976} & .908 & .333 & .111 & .524 & .593 & .857 & .883 \\ 
        BERTScore$^{\text{R}}$ & .857 & .889 & .119 & .115 & .310 & .604 & .714 & .860 \\ \hline
        \multicolumn{9}{c}{Distribution-Based Metrics} \\ \hline
        FBD$^{\text{B}}$ & .762 & .790 & .190 & .203 & \textbf{.667} & .404 & .786 & .763\\
        FBD$^{\text{R}}$ & \textbf{.976} & \textbf{.965} & \textbf{.429} & \textbf{.472} & .524 & \textbf{.839} & \textbf{.929} & \textbf{.963} \\
        PRD$^{\text{B}}$ & .833 & .932 & .238 & .156 & .571 & .708 & .690 & .914  \\ 
        PRD$^{\text{R}}$ & .881 & .925 & .238 & .235 & .452 & .729 & .762 & .906 \\ \bottomrule
	\end{tabularx}
	\centering
	\begin{tablenotes}
    \small
    \item 1.``Spr." and ``Pr." refer to Spearman and Pearson correlation coefficients, respectively.
    \item 2. $^{\text{B}}$ and $^{\text{R}}$ mean using BERT (base) and RoBERTa (base) as language models, respectively.
\end{tablenotes}
	\caption{\label{tab:fine-grained} Fine-grained comparisons on Daily(Z)~\cite{zhao2020designing} corpus.
	}
	\label{tab:slerp}
\end{table} 
Surprisingly, RoBERTa-based metrics, including BERTScore, the proposed FBD and PRD, perform better than the corresponding BERT-base ones. 
Given that our FBD metric lies on the assumption of multivariate Gaussian distribution, we hypothesize that the semantic representations extracted by RoBERTa model fit Gaussian distribution better than BERT model. To verify this point, as shown in Table~\ref{normality-table}, we use ``Shapiro–Wilk test"~\cite{razali2011power} to calculate the normality in statistics, where small values lead to the rejection of normality whereas a value of one indicates normality of the data.

\subsection{Fine-Grained Performances} 

In the evaluation corpus Daily(Z)~\cite{zhao2020designing}, it provides four fine-grained human evaluation scores, including \textit{relevance}, \textit{grammar}, \textit{content} and \textit{overall}, which can be used to dive more insights of different evaluation metrics.

As shown in Table~\ref{tab:fine-grained}, our proposed metric FBD$^{\text{R}}$ achieves the best performances on most evaluations in the fine-grained comparisons. It indicates the distribution-wise metric correlate better with human judgements on various aspects.

\section{Conclusions}
In this paper, we propose to measure the performance of a dialogue system by computing the distribution-wise difference between its generated conversations and real-world conversations. Specifically, two distribution-wise metrics, FBD and PRD, are developed on pre-trained language models. Experiments on six public dialogue corpora show that our proposed metrics correlate better with human judgments than existing metrics. 

\clearpage
%\balance
\bibliographystyle{acl_natbib}
\bibliography{acl2021}

\begin{thebibliography}{39}
\expandafter\ifx\csname natexlab\endcsname\relax\def\natexlab#1{#1}\fi

\bibitem[{Banerjee and Lavie(2005)}]{Banerjee2005METEORAA}
Satanjeev Banerjee and Alon Lavie. 2005.
\newblock Meteor: An automatic metric for mt evaluation with improved
  correlation with human judgments.
\newblock In \emph{Proceedings of the ACL workshop on intrinsic and extrinsic
  evaluation measures for machine translation and/or summarization}.

\bibitem[{Chan et~al.(2021)Chan, Liu, Li, Zhang, Zhao, Shi, and
  Yan}]{chan2021enhancing}
Zhangming Chan, Lemao Liu, Juntao Li, Haisong Zhang, Dongyan Zhao, Shuming Shi,
  and Rui Yan. 2021.
\newblock Enhancing the open-domain dialogue evaluation in latent space.
\newblock In \emph{Findings of ACL}.

\bibitem[{Chen et~al.(2021)Chen, Li, and King}]{wang2021a}
Wang Chen, Piji Li, and Irwin King. 2021.
\newblock A training-free and reference-free summarization evaluation metric
  via centrality-weighted relevance and self-referenced redundancy.
\newblock In \emph{ACL}.

\bibitem[{Clark et~al.(2020)Clark, Luong, Le, and Manning}]{clark2020electra}
Kevin Clark, Minh-Thang Luong, Quoc~V Le, and Christopher~D Manning. 2020.
\newblock Electra: Pre-training text encoders as discriminators rather than
  generators.
\newblock In \emph{ICLR}.

\bibitem[{Denkowski and Lavie(2014)}]{denkowski2014meteor}
Michael Denkowski and Alon Lavie. 2014.
\newblock Meteor universal: Language specific translation evaluation for any
  target language.
\newblock In \emph{Proceedings of the workshop on statistical machine
  translation}.

\bibitem[{Devlin et~al.(2019)Devlin, Chang, Lee, and Toutanova}]{jacob2018bert}
Jacob Devlin, Ming{-}Wei Chang, Kenton Lee, and Kristina Toutanova. 2019.
\newblock Bert: Pre-training of deep bidirectional transformers for language
  understanding.
\newblock In \emph{NAACL}.

\bibitem[{Dinan et~al.(2019)Dinan, Logacheva, Malykh, Miller, Shuster, Urbanek,
  Kiela, Szlam, Serban, Lowe et~al.}]{dinan2019second}
Emily Dinan, Varvara Logacheva, Valentin Malykh, Alexander Miller, Kurt
  Shuster, Jack Urbanek, Douwe Kiela, Arthur Szlam, Iulian Serban, Ryan Lowe,
  et~al. 2019.
\newblock The second conversational intelligence challenge (convai2).
\newblock \emph{arXiv preprint arXiv:1902.00098}.

\bibitem[{Dowson and Landau(1982)}]{dowson1982frechet}
DC~Dowson and BV~Landau. 1982.
\newblock The fr{\'e}chet distance between multivariate normal distributions.
\newblock \emph{Journal of multivariate analysis}, 12(3):450--455.

\bibitem[{Foltz et~al.(1998)Foltz, Kintsch, and Landauer}]{Foltz1998TheMO}
Peter~W Foltz, Walter Kintsch, and Thomas~K Landauer. 1998.
\newblock The measurement of textual coherence with latent semantic analysis.
\newblock \emph{Discourse processes}, 25(2-3):285--307.

\bibitem[{Forgues et~al.(2014)Forgues, Pineau, Larchev{\^e}que, and
  Tremblay}]{forgues2014bootstrapping}
Gabriel Forgues, Joelle Pineau, Jean-Marie Larchev{\^e}que, and R{\'e}al
  Tremblay. 2014.
\newblock Bootstrapping dialog systems with word embeddings.
\newblock In \emph{NeurIPS, modern machine learning and natural language
  processing workshop}.

\bibitem[{Gao et~al.(2021)Gao, Bi, Xu, and Shi}]{gao2021ream}
Jun Gao, Wei Bi, Ruifeng Xu, and Shuming Shi. 2021.
\newblock $\mbox{REAM}\sharp$: An enhancement approach to reference-based
  evaluation metrics for open-domain dialog generation.
\newblock In \emph{Findings of ACL}.

\bibitem[{Ghazarian et~al.(2019)Ghazarian, Wei, Galstyan, and
  Peng}]{Ghazarian2019BetterAE}
Sarik Ghazarian, Johnny Tian-Zheng Wei, A.~Galstyan, and Nanyun Peng. 2019.
\newblock Better automatic evaluation of open-domain dialogue systems with
  contextualized embeddings.
\newblock \emph{ArXiv}, abs/1904.10635.

\bibitem[{Heusel et~al.(2017)Heusel, Ramsauer, Unterthiner, Nessler, and
  Hochreiter}]{heusel2017gans}
Martin Heusel, Hubert Ramsauer, Thomas Unterthiner, Bernhard Nessler, and Sepp
  Hochreiter. 2017.
\newblock Gans trained by a two time-scale update rule converge to a local nash
  equilibrium.
\newblock In \emph{NeurIPS}.

\bibitem[{Huang et~al.(2020)Huang, Ye, Qin, Lin, and Liang}]{huang2020grade}
Lishan Huang, Zheng Ye, Jinghui Qin, Liang Lin, and Xiaodan Liang. 2020.
\newblock Grade: Automatic graph-enhanced coherence metric for evaluating
  open-domain dialogue systems.
\newblock In \emph{EMNLP}.

\bibitem[{Karras et~al.(2017)Karras, Aila, Laine, and
  Lehtinen}]{karras2017progressive}
Tero Karras, Timo Aila, Samuli Laine, and Jaakko Lehtinen. 2017.
\newblock Progressive growing of gans for improved quality, stability, and
  variation.
\newblock \emph{arXiv preprint arXiv:1710.10196}.

\bibitem[{Lin(2004)}]{Lin2004ROUGEAP}
Chin-Yew Lin. 2004.
\newblock Rouge: A package for automatic evaluation of summaries.
\newblock In \emph{ACL}.

\bibitem[{Liu et~al.(2018)Liu, Bi, Gao, Liu, Yao, and Shi}]{liu2018towards}
Yahui Liu, Wei Bi, Jun Gao, Xiaojiang Liu, Jian Yao, and Shuming Shi. 2018.
\newblock Towards less generic responses in neural conversation models: A
  statistical re-weighting method.
\newblock In \emph{EMNLP}.

\bibitem[{Liu et~al.(2019)Liu, Ott, Goyal, Du, Joshi, Chen, Levy, Lewis,
  Zettlemoyer, and Stoyanov}]{liu2019roberta}
Yinhan Liu, Myle Ott, Naman Goyal, Jingfei Du, Mandar Joshi, Danqi Chen, Omer
  Levy, Mike Lewis, Luke Zettlemoyer, and Veselin Stoyanov. 2019.
\newblock Roberta: A robustly optimized bert pretraining approach.
\newblock \emph{arXiv preprint arXiv:1907.11692}.

\bibitem[{Lowe et~al.(2017)Lowe, Noseworthy, Serban, Angelard-Gontier, Bengio,
  and Pineau}]{Lowe2017TowardsAA}
Ryan Lowe, Michael Noseworthy, I.~Serban, Nicolas Angelard-Gontier, Yoshua
  Bengio, and Joelle Pineau. 2017.
\newblock Towards an automatic turing test: Learning to evaluate dialogue
  responses.
\newblock In \emph{ACL}.

\bibitem[{Mehri and Esk{\'e}nazi(2020)}]{Mehri2020USRAU}
Shikib Mehri and M.~Esk{\'e}nazi. 2020.
\newblock Usr: An unsupervised and reference free evaluation metric for dialog
  generation.
\newblock In \emph{ACL}.

\bibitem[{Miller et~al.(2017)Miller, Feng, Fisch, Lu, Batra, Bordes, Parikh,
  and Weston}]{miller2017parlai}
Alexander~H Miller, Will Feng, Adam Fisch, Jiasen Lu, Dhruv Batra, Antoine
  Bordes, Devi Parikh, and Jason Weston. 2017.
\newblock Parlai: A dialog research software platform.
\newblock \emph{arXiv preprint arXiv:1705.06476}.

\bibitem[{Mitchell and Lapata(2008)}]{Mitchell2008VectorbasedMO}
Jeff Mitchell and Mirella Lapata. 2008.
\newblock Vector-based models of semantic composition.
\newblock In \emph{ACL}.

\bibitem[{Papineni et~al.(2002)Papineni, Roukos, Ward, and
  Zhu}]{Papineni2002BleuAM}
Kishore Papineni, S.~Roukos, T.~Ward, and Wei-Jing Zhu. 2002.
\newblock Bleu: a method for automatic evaluation of machine translation.
\newblock In \emph{ACL}.

\bibitem[{Park et~al.(2019)Park, Liu, Wang, and Zhu}]{park2019semantic}
Taesung Park, Ming-Yu Liu, Ting-Chun Wang, and Jun-Yan Zhu. 2019.
\newblock Semantic image synthesis with spatially-adaptive normalization.
\newblock In \emph{CVPR}.

\bibitem[{Rashkin et~al.(2018)Rashkin, Smith, Li, and
  Boureau}]{rashkin2018towards}
Hannah Rashkin, Eric~Michael Smith, Margaret Li, and Y-Lan Boureau. 2018.
\newblock Towards empathetic open-domain conversation models: A new benchmark
  and dataset.
\newblock \emph{arXiv preprint arXiv:1811.00207}.

\bibitem[{Razali et~al.(2011)Razali, Wah et~al.}]{razali2011power}
Nornadiah~Mohd Razali, Yap~Bee Wah, et~al. 2011.
\newblock Power comparisons of shapiro-wilk, kolmogorov-smirnov, lilliefors and
  anderson-darling tests.
\newblock \emph{Journal of statistical modeling and analytics}, 2(1):21--33.

\bibitem[{Rus and Lintean(2012)}]{rus2012optimal}
Vasile Rus and Mihai Lintean. 2012.
\newblock An optimal assessment of natural language student input using
  word-to-word similarity metrics.
\newblock In \emph{International Conference on Intelligent Tutoring Systems}.
  Springer.

\bibitem[{Sajjadi et~al.(2018)Sajjadi, Bachem, Lucic, Bousquet, and
  Gelly}]{sajjadi2018assessing}
Mehdi~SM Sajjadi, Olivier Bachem, Mario Lucic, Olivier Bousquet, and Sylvain
  Gelly. 2018.
\newblock Assessing generative models via precision and recall.
\newblock In \emph{NeurIPS}.

\bibitem[{Sellam et~al.(2020)Sellam, Das, and Parikh}]{sellam2020bleurt}
Thibault Sellam, Dipanjan Das, and Ankur Parikh. 2020.
\newblock Bleurt: Learning robust metrics for text generation.
\newblock In \emph{ACL}.

\bibitem[{Tao et~al.(2018)Tao, Mou, Zhao, and Yan}]{tao2018ruber}
Chongyang Tao, Lili Mou, Dongyan Zhao, and Rui Yan. 2018.
\newblock Ruber: An unsupervised method for automatic evaluation of open-domain
  dialog systems.
\newblock In \emph{AAAI}.

\bibitem[{Vinyals and Le(2015)}]{vinyals2015neural}
Oriol Vinyals and Quoc Le. 2015.
\newblock A neural conversational model.
\newblock \emph{arXiv preprint arXiv:1506.05869}.

\bibitem[{Wieting et~al.(2015)Wieting, Bansal, Gimpel, and
  Livescu}]{wieting2015towards}
John Wieting, Mohit Bansal, Kevin Gimpel, and Karen Livescu. 2015.
\newblock Towards universal paraphrastic sentence embeddings.
\newblock \emph{arXiv preprint arXiv:1511.08198}.

\bibitem[{Yang et~al.(2019)Yang, Dai, Yang, Carbonell, Salakhutdinov, and
  Le}]{yang2019xlnet}
Zhilin Yang, Zihang Dai, Yiming Yang, Jaime Carbonell, Russ~R Salakhutdinov,
  and Quoc~V Le. 2019.
\newblock Xlnet: Generalized autoregressive pretraining for language
  understanding.
\newblock In \emph{NeurIPS}.

\bibitem[{Zhang et~al.(2018{\natexlab{a}})Zhang, Xu, Li, Zhang, Wang, Huang,
  and Metaxas}]{zhang2018stackgan++}
Han Zhang, Tao Xu, Hongsheng Li, Shaoting Zhang, Xiaogang Wang, Xiaolei Huang,
  and Dimitris~N Metaxas. 2018{\natexlab{a}}.
\newblock Stackgan++: Realistic image synthesis with stacked generative
  adversarial networks.
\newblock \emph{IEEE Transactions on Pattern Analysis and Machine Intelligence
  (TPAMI)}, 41(8):1947--1962.

\bibitem[{Zhang et~al.(2018{\natexlab{b}})Zhang, Dinan, Urbanek, Szlam, Kiela,
  and Weston}]{zhang2018personalizing}
Saizheng Zhang, Emily Dinan, Jack Urbanek, Arthur Szlam, Douwe Kiela, and Jason
  Weston. 2018{\natexlab{b}}.
\newblock Personalizing dialogue agents: I have a dog, do you have pets too?
\newblock \emph{arXiv preprint arXiv:1801.07243}.

\bibitem[{Zhang et~al.(2020)Zhang, Kishore, Wu, Weinberger, and
  Artzi}]{zhang2019bertscore}
Tianyi Zhang, Varsha Kishore, Felix Wu, Kilian~Q Weinberger, and Yoav Artzi.
  2020.
\newblock Bertscore: Evaluating text generation with bert.
\newblock In \emph{ICLR}.

\bibitem[{Zhao et~al.(2017)Zhao, Zhao, and Eskenazi}]{zhao-etal-2017-learning}
Tiancheng Zhao, Ran Zhao, and Maxine Eskenazi. 2017.
\newblock Learning discourse-level diversity for neural dialog models using
  conditional variational autoencoders.
\newblock In \emph{ACL}.

\bibitem[{Zhao et~al.(2020)Zhao, Lala, and Kawahara}]{zhao2020designing}
Tianyu Zhao, Divesh Lala, and Tatsuya Kawahara. 2020.
\newblock Designing precise and robust dialogue response evaluators.
\newblock In \emph{ACL}.

\bibitem[{Zhou et~al.(2017)Zhou, Luo, Cao, Lin, Chen, and
  He}]{zhou2017mechanism}
Ganbin Zhou, Ping Luo, Rongyu Cao, Fen Lin, Bo~Chen, and Qing He. 2017.
\newblock Mechanism-aware neural machine for dialogue response generation.
\newblock In \emph{AAAI}.

\end{thebibliography}

\appendix

\end{document}